\documentstyle[twocolumn]{article}
\newlength{\colingsep}
\font\fr=eufm10
\newcounter{propcounter}
\long\def\prop[#1 \par#2\prop]
{%
\refstepcounter{propcounter}%
{%
\parindent 0pt%
\parskip 0pt%
\par%
\bf#1 \thepropcounter. \sl#2%
}%
\par\noindent%
}
\long\def\proof[ #1\proof]
{%
{%
\parindent 0pt%
\parskip 0pt%
\par%
\bf Proof. \rm #1\rule{5pt}{5pt}%
}%
\par\noindent%
}
\long\def\disp[#1\disp]%
{
\begin{list}
{}
{
\parskip 1pt
\parsep 1pt
\partopsep 0pt
\topsep 0pt
\itemsep 0pt
\labelsep 0pt
\labelwidth 8pt
\leftmargin \labelwidth
\rightmargin 0pt
}
\item
\raggedright
#1
\end{list}
}
\long\def\refs[#1\refs]%
{
\begin{list}
{}
{
\parskip 1pt
\parsep 1pt
\partopsep 0pt
\topsep 0pt
\itemsep 0pt
\labelsep 8pt
\labelwidth 0pt
\leftmargin \labelsep
\rightmargin 0pt
}
#1
\end{list}
}
\long\def\ind[#1\ind=#2\ind]%
{
\begin{list}
{#1}
{
\parskip 1pt
\parsep 1pt
\partopsep 0pt
\topsep 0pt
\itemsep 0pt
\labelsep 0pt
\settowidth{\labelwidth}{#1}
\leftmargin \labelwidth
\rightmargin 0pt
}
\item
#2
\end{list}
}
\long\def\imp[#1\imp]%
{
\ind[$\Longrightarrow$ \ind=#1\ind]
}
\long\def\ifthen[ #1 #2 then #3 \ifthen]%
{
\ind[#1 \ind=#2\ind]
\ind[then \ind=#3\ind]
}
\long\def\ifandonlyif[ #1 iff #2 \ifandonlyif]%
{
#1
\ind[iff \ind= #2\ind]
}
\long\def\ifthenotherwise[ #1 #2 then #3 otherwise #4 \ifthenotherwise]%
{
\ind[#1 \ind=#2\ind]
\ind[then \ind=#3\ind]
\ind[otherwise \ind=#4\ind]
}
\def\cite#1{{\sc[#1]}}
\def\sequence#1{\langle#1\rangle}
\def\N{\mbox{\rm I\hspace{-.5mm}N}}
\def\Q{\mbox{\fr Q}}
\def\T{\mbox{\fr T}}
\def\S{\mbox{\fr S}}
\def\F{\mbox{\fr F}}
\def\A{\mbox{\fr A}}
\def\P{\mbox{\fr P}}
\def\Epsilon{\mbox{\rm E}}
\def\UU{\widetilde{U}}
\def\SS{\widetilde{S}}
\def\AA{\widetilde{A}}
\def\II{\widetilde{I}}

\def\App{\mbox{\tt APP}}
\def\Sub{\mbox{\tt SUB}}
\def\Res{\mbox{\tt RES}}
\def\Gen{\mbox{\tt GEN}}
\def\Test{\mbox{\tt TEST}}
\def\Listin{\mbox{$\Sigma_{\mbox{\scriptsize\rm in}}$}}
\def\Listout{\mbox{$\Sigma_{\mbox{\scriptsize\rm out}}$}}
\def\Sat{\mbox{\tt SAT}}

\title{\vskip -10pt
{\begin{minipage}[b]{5in}
\centerline{TYPED FEATURE STRUCTURES AS DESCRIPTIONS}
\vskip 14pt
\centerline{Paul John King\thanks{The research presented in this paper
was sponsored by Teilprojekt B4 ``Constraints on Grammar for Efficient
Generation'' of the Sonderforschungsbereich 340 of the Deutsche
Forschungsgemeinschaft. I also wish to thank Bob Carpenter, Dale
Gerdemann, Thilo G\"otz and Jennifer King for their invaluable help
with this paper.}}
\vskip 14pt
\centerline{Seminar f\"ur Sprachwissenschaft,
Eberhard-Karls-Universit\"at\thanks{Wilhelmstr.~113, 72074 T\"ubingen,
Germany.  Email: king@sfs.nphil.uni-tuebingen.de.}}
\end{minipage}}
\vskip -30pt}
\author{}

\begin{document}

\maketitle\thispagestyle{empty}

\subsection*{\raggedright ABSTRACT}

A description is an entity that can be interpreted as true or false of
an object, and using feature structures as descriptions accrues
several computational benefits.  In this paper, I create an explicit
interpretation of a typed feature structure used as a description,
define the notion of a satisfiable feature structure, and create a
simple and effective algorithm to decide if a feature structure is
satisfiable.

\subsection*{\raggedright 1. INTRODUCTION}
Describing objects is one of several purposes for which linguists use
feature structures.  A description is an entity that can be
interpreted as true or false of an object.  For example, the
conventional interpretation of the description `it is black' is true
of a soot particle, but false of a snowflake.  Therefore, any use of a
feature structure to describe an object demands that the feature
structure can be interpreted as true or false of the object.  In this
paper, I tailor the semantics of {\cite{King 1989}} to suit the typed
feature structures of {\cite{Carpenter 1992}}, and so create an
explicit interpretation of a typed feature structure used as a
description.  I then use this interpretation to define the notion of a
satisfiable feature structure.

Though no feature structure algebra provides descriptions as
expressive as those provided by a feature logic, using feature
structures to describe objects profits from a large stock of available
computational techniques to represent, test and process feature
structures.  In this paper, I demonstrate the computational benefits
of marrying a tractable syntax and an explicit semantics by creating a
simple and effective algorithm to decide the satisfiability of a
feature structure.  Gerdemann and G\"otz's Troll type resolution
system implements both the semantics and an efficient refinement of
the satisfiability algorithm I present here (see {\cite{G\"otz 1993}},
{\cite{Gerdemann and King 1994}} and {\cite{Gerdemann (fc)}}).

\subsection*{\raggedright 2. A FEATURE STRUCTURE SEMANTICS}

A signature provides the symbols from which to construct typed feature
structures, and an interpretation gives those symbols meaning.
\prop[
Definition

$\Sigma$ is a signature iff
\disp[
$\Sigma$ is a sextuple $\sequence{\Q,\T,\preceq,\S,\A,\F}$,

$\Q$ is a set,

$\sequence{\T,\preceq}$ is a partial order,

$\S=
\left\{
\sigma\in\T
\left|
\begin{tabular}{l}
for each $\tau\in\T$,
\\
\hspace{4pt}if $\sigma\preceq\tau$ then $\sigma=\tau$
\end{tabular}
\right.
\right\}$,

$\A$ is a set,

$\F$ is a partial function from the Cartesian product of $\T$ and $\A$
to $\T$, and

for
each $\tau\in\T$,
each $\tau'\in\T$ and
each $\alpha\in\A$,%
\disp[
\ifthen[
if
$\F(\tau,\alpha)$ is defined and
$\tau\preceq\tau'$
then
$\F(\tau',\alpha)$ is defined, and
\\
$\F(\tau,\alpha)\preceq\F(\tau',\alpha)$.
\ifthen]
\disp]
\disp]
\prop]
Henceforth, I tacitly work with a signature
$\sequence{\Q,\T,\preceq,\S,\A,\F}$.  I call members of $\Q$ states,
members of $\T$ types, $\preceq$ subsumption, members of $\S$ species,
members of $\A$ attributes, and $\F$ appropriateness.
\prop[
Definition

$I$ is an interpretation iff
\disp[
$I$ is a triple $\sequence{U,S,A}$,

$U$ is a set,

$S$ is a total function from $U$ to $\S$

$A$ is a total function from $\A$ to the set of partial functions from
$U$ to $U$,

for
each $\alpha\in\A$ and
each $u\in U$,
\disp[
\ifthen[
if
$A(\alpha)(u)$ is defined
then
$\F(S(u),\alpha)$ is defined, and
\\
$\F(S(u),\alpha)\preceq S(A(\alpha)(u))$, and
\ifthen]
\disp]

for
each $\alpha\in\A$ and
each $u\in U$,
\disp[
\ifthen[
if
$\F(S(u),\alpha)$ is defined
then
$A(\alpha)(u)$ is defined.
\ifthen]
\disp]
\disp]
\prop]
Suppose that $I$ is an interpretation $\sequence{U,S,A}$.  I call each
member of $U$ an object in $I$.

Each type denotes a set of objects in $I$.  The denotations of the
species partition $U$, and $S$ assigns each object in $I$ the unique
species whose denotation contains the object: object $u$ is in the
denotation of species $\sigma$ iff $\sigma=S(u)$.  Subsumption encodes
a relationship between the denotations of species and types: object
$u$ is in the denotation of type $\tau$ iff $\tau\preceq S(u)$.  So,
if $\tau_1\preceq\tau_2$ then the denotation of type $\tau_1$ contains
the denotation of type $\tau_2$.

Each attribute denotes a partial function from the objects in $I$ to
the objects in $I$, and $A$ assigns each attribute the partial
function it denotes.  Appropriateness encodes a relationship between
the denotations of species and attributes: if
$\F\sequence{\sigma,\alpha}$ is defined then the denotation of
attribute $\alpha$ acts upon each object in the denotation of species
$\sigma$ to yield an object in the denotation of type
$\F\sequence{\sigma,\alpha}$, but if $\F\sequence{\sigma,\alpha}$ is
undefined then the denotation of attribute $\alpha$ acts upon no
object in the denotation of species $\sigma$.  So, if
$\F\sequence{\tau,\alpha}$ is defined then the denotation of attribute
$\alpha$ acts upon each object in the denotation of type $\tau$ to
yield an object in the denotation of type $\F\sequence{\tau,\alpha}$.

I call a finite sequence of attributes a path, and write $\P$ for the
set of paths.
\prop[
Definition

$P$ is the path interpretation function under $I$ iff
\disp[
$I$ is an interpretation $\sequence{U,S,A}$,

$P$ is a total function from $\P$ to the set of partial functions from
$U$ to $U$, and

for
each $\sequence{\alpha_1,\ldots,\alpha_n}\in\P$,
\disp[
$P\sequence{\alpha_1,\ldots,\alpha_n}$ is the functional composition
of $A(\alpha_1),\ldots,A(\alpha_n)$.
\disp]
\disp]
\prop]
I write $P_I$ for the path interpretation function under $I$.

\prop[
Definition

$F$ is a feature structure iff
\disp[
$F$ is a quadruple $\sequence{Q,q,\delta,\theta}$,

$Q$ is a finite subset of $\Q$,

$q\in Q$,

$\delta$ is a finite partial function from the Cartesian product of
$Q$ and $\A$ to $Q$,

$\theta$ is a total function from $Q$ to $\T$, and

for
each $q'\in Q$,
\disp[
for
some $\pi\in\P$,
$\pi$ runs to $q'$ in $F$,
\disp]
\disp]
where
$\sequence{\alpha_1,\ldots,\alpha_n}$ runs to $q'$ in $F$ iff
\disp[
$\sequence{\alpha_1,\ldots,\alpha_n}\in\P$,

$q'\in Q$, and

for
some $\{q_0,\ldots,q_n\}\subseteq Q$,
\disp[
$q=q_0$,

for
each $i<n$,
\disp[
$\delta(q_i,\alpha_{i+1})$ is defined, and
\\
$\delta(q_i,\alpha_{i+1})=q_{i+1}$, and
\disp]

$q_n=q'$.
\disp]
\disp]
\prop]
Each feature structure is a connected Moore machine (see {\cite{Moore
1956}}) with finitely many states, input alphabet $\A$, and output
alphabet\- $\T$.

\prop[
Definition

$F$ is true of $u$ under $I$ iff
\disp[
$F$ is a feature structure $\sequence{Q,q,\delta,\theta}$,

$I$ is an interpretation $\sequence{U,S,A}$,

$u$ is an object in $I$, and

for
each $\pi_1\in\P$,
each $\pi_2\in\P$ and
each $q'\in Q$,
\disp[
\ifthen[
if
$\pi_1$ runs to $q'$ in $F$, and
\\
$\pi_2$ runs to $q'$ in $F$
then
$P_I(\pi_1)(u)$ is defined,
\\
$P_I(\pi_2)(u)$ is defined,
\\
$P_I(\pi_1)(u)=P_I(\pi_2)(u)$, and
\\
$\theta(q')\preceq S(P_I(\pi_1)(u))$.
\ifthen]
\disp]
\disp]
\prop]
\prop[
Definition

$F$ is a satisfiable feature structure iff
\disp[
$F$ is a feature structure, and

for
some interpretation $I$ and
some object $u$ in $I$,
$F$ is true of $u$ under $I$.
\disp]
\prop]

\subsection*{\raggedright 3. MORPHS}

The abundance of interpretations seems to preclude an effective
algorithm to decide if a feature structure is satisfiable.  However, I
insert {\sl morphs} between feature structures and objects to yield an
interpretation free characterisation of a satisfiable feature
structure.

\prop[
Definition

$M$ is a semi-morph iff
\disp[
$M$ is a triple $\sequence{\Delta,\Gamma,\Lambda}$,

$\Delta$ is a nonempty subset of $\P$,

$\Gamma$ is an equivalence relation over $\Delta$,

for
each $\alpha\in\A$,
each $\pi_1\in\P$ and
each $\pi_2\in\P$,
\disp[
\ifthen[
if
$\pi_1\alpha\in\Delta$ and
$\sequence{\pi_1,\pi_2}\in\Gamma$
then
$\sequence{\pi_1\alpha,\pi_2\alpha}\in\Gamma$,
\ifthen]
\disp]

$\Lambda$ is a total function from $\Delta$ to $\S$,

for
each $\pi_1\in\P$ and
each $\pi_2\in\P$,
\disp[
if
$\sequence{\pi_1,\pi_2}\in\Gamma$
then
$\Lambda(\pi_1)=\Lambda(\pi_2)$, and
\disp]

for
each $\alpha\in\A$ and
each $\pi\in\P$,
\disp[
\ifthen[
if
$\pi\alpha\in\Delta$
then
$\pi\in\Delta$,
$\F(\Lambda(\pi),\alpha)$ is defined, and
\\
$\F(\Lambda(\pi),\alpha)\preceq\Lambda(\pi\alpha)$.
\ifthen]
\disp]
\disp]
\prop]
\prop[
Definition

$M$ is a morph iff
\disp[
$M$ is a semi-morph $\sequence{\Delta,\Gamma,\Lambda}$, and

for
each $\alpha\in\A$ and
each $\pi\in\P$,
\disp[
\ifthen[
if
$\pi\in\Delta$ and
$\F(\Lambda(\pi),\alpha)$ is defined
then
$\pi\alpha\in\Delta$.
\ifthen]
\disp]
\disp]
\prop]
Each morph is the Moshier abstraction (see {\cite{Moshier 1988}}) of a
connected and totally well-typed (see {\cite{Carpenter 1992}}) Moore
machine with possibly infinitely many states, input alphabet $\A$, and
output alphabet $\S$.

\pagebreak

\prop[
Definition

$M$ abstracts $u$ under $I$ iff
\disp[
$M$ is a morph $\sequence{\Delta,\Gamma,\Lambda}$,

$I$ is an interpretation $\sequence{U,S,A}$,

$u$ is an object in $I$,

for
each $\pi_1\in\P$ and
each $\pi_2\in\P$,
\disp[
\ifandonlyif[
$\sequence{\pi_1,\pi_2}\in\Gamma$
iff
$P_I(\pi_1)(u)$ is defined,
\\
$P_I(\pi_2)(u)$ is defined, and
\\
$P_I(\pi_1)(u)=P_I(\pi_2)(u)$, and
\ifandonlyif]
\disp]

for
each $\sigma\in\S$ and
each $\pi\in\P$,
\disp[
\ifandonlyif[
$\sequence{\pi,\sigma}\in\Lambda$
iff
$P_I(\pi)(u)$ is defined, and
\\
$\sigma=S(P_I(\pi)(u))$.
\ifandonlyif]
\disp]
\disp]
\prop]
\prop[
Proposition

For
each interpretation $I$ and
each object $u$ in $I$,
\disp[
some unique morph abstracts $u$ under $I$.
\disp]
\prop]
I thus write of {\sl the} abstraction of $u$ under $I$.

\prop[
Definition

$u$ is a standard object iff
\disp[
$u$ is a quadruple $\sequence{\Delta,\Gamma,\Lambda,\Epsilon}$,

$\sequence{\Delta,\Gamma,\Lambda}$ is a morph, and

$\Epsilon$ is an equivalence class under $\Gamma$.
\disp]
\prop]
I write $\UU$ for the set of standard objects, write $\SS$ for the
total function from $\UU$ to $\S$, where
\disp[
for
each $\sigma\in\S$ and
each $\sequence{\Delta,\Gamma,\Lambda,\Epsilon}\in\UU$,
\disp[
\ifandonlyif[
$\SS\sequence{\Delta,\Gamma,\Lambda,\Epsilon}=\sigma$
iff
for
some $\pi\in\Epsilon$,
$\Lambda(\pi)=\sigma$,
\ifandonlyif]
\disp]
\disp]
and write $\AA$ for the total function from $\A$ to the set of partial
functions from $\UU$ to $\UU$, where
\disp[
for
each $\alpha\in\A$,
each $\sequence{\Delta,\Gamma,\Lambda,\Epsilon}\in\UU$ and
each $\sequence{\Delta',\Gamma',\Lambda',\Epsilon'}\in\UU$,
\disp[
\ifandonlyif[
$\AA(\alpha)\sequence{\Delta,\Gamma,\Lambda,\Epsilon}$ is defined, and
\\
$\AA(\alpha)\sequence{\Delta,\Gamma,\Lambda,\Epsilon}
=\sequence{\Delta',\Gamma',\Lambda',\Epsilon'}$
iff
$\sequence{\Delta,\Gamma,\Lambda}=\sequence{\Delta',\Gamma',\Lambda'}$,
and
\\
for
some $\pi\in\Epsilon$,
$\pi\alpha\in\Epsilon'$.
\ifandonlyif]
\disp]
\disp]
\prop[
Lemma

$\sequence{\UU,\SS,\AA}$ is an interpretation.
\prop]
I write $\II$ for $\sequence{\UU,\SS,\AA}$.
\prop[
Lemma

For
each $\sequence{\Delta,\Gamma,\Lambda,\Epsilon}\in\UU$,
each $\sequence{\Delta',\Gamma',\Lambda',\Epsilon'}\in\UU$ and
each $\pi\in\P$,
\disp[
\ifandonlyif[
$P_{\II}(\pi)\sequence{\Delta,\Gamma,\Lambda,\Epsilon}$ is defined, and
\\
$P_{\II}(\pi)\sequence{\Delta,\Gamma,\Lambda,\Epsilon}
=\sequence{\Delta',\Gamma',\Lambda',\Epsilon'}$
iff
$\sequence{\Delta,\Gamma,\Lambda}=\sequence{\Delta',\Gamma',\Lambda'}$,
and
\\
for
some $\pi'\in\Epsilon$,
$\pi'\pi\in\Epsilon'$.
\ifandonlyif]
\disp]
\prop]
\proof[
By induction on the length of $\pi$.
\proof]
\prop[
Lemma

For
each $\sequence{\Delta,\Gamma,\Lambda,\Epsilon}\in\UU$,
\disp[
\ifthen[
if
$\Epsilon$ is the equivalence class of the empty path under $\Gamma$
then
the abstraction of $\sequence{\Delta,\Gamma,\Lambda,\Epsilon}$ under
$\II$ is $\sequence{\Delta,\Gamma,\Lambda}$.
\ifthen]
\disp]
\prop]
\prop[
Proposition
\label{M_is_the_abstraction_of_some_u}

For
each morph $M$,
\disp[
for
some interpretation $I$ and
some object $u$ in $I$,
\disp[
$M$ is the abstraction of $u$ under $I$.
\disp]
\disp]
\prop]

\pagebreak

\prop[
Definition

$F$ approximates $M$ iff
\disp[
$F$ is a feature structure $\sequence{Q,q,\delta,\theta}$,

$M$ is a morph $\sequence{\Delta,\Gamma,\Lambda}$, and

for
each $\pi_1\in\P$,
each $\pi_2\in\P$ and
each $q'\in Q$,
\disp[
\ifthen[
if
$\pi_1$ runs to $q'$ in $F$, and
\\
$\pi_2$ runs to $q'$ in $F$
then
$\sequence{\pi_1,\pi_2}\in\Gamma$, and
\\
$\theta(q')\preceq\Lambda(\pi_1)$.
\ifthen]
\disp]
\disp]
\prop]
A feature structure approximates a morph iff the Moshier abstraction
of the feature structure abstractly subsumes (see {\cite{Carpenter
1992}}) the morph.
\prop[
Proposition
\label{F_is_true_of_u_iff_F_approximates_the_abstraction_of_u}

For
each interpretation $I$,
each object $u$ in $I$ and
each feature structure $F$,
\disp[
\ifandonlyif[
$F$ is true of $u$ under $I$
iff
$F$ approximates the abstraction of $u$ under $I$.
\ifandonlyif]
\disp]
\prop]

\prop[
Theorem
\label{F_is_satisfiable_iff_F_approximates_some_morph}

For each feature structure $F$,
\disp[
$F$ is satisfiable
iff
$F$ approximates some morph.
\disp]
\prop]
\proof[
{}From
propositions~\ref{M_is_the_abstraction_of_some_u}
and~\ref{F_is_true_of_u_iff_F_approximates_the_abstraction_of_u}.
\proof]

\subsection*{\raggedright 4. RESOLVED FEATURE STRUCTURES}

Though theorem~\ref{F_is_satisfiable_iff_F_approximates_some_morph}
gives an interpretation free characterisation of a satisfiable feature
structure, the characterisation still seems to admit of no effective
algorithm to decide if a feature structure is satisfiable.  However, I
use theorem~\ref{F_is_satisfiable_iff_F_approximates_some_morph} and
{\sl resolved feature structures} to yield a less general
interpretation free characterisation of a satisfiable feature
structure that admits of such an algorithm.

\prop[
Definition

$R$ is a resolved feature structure iff
\disp[
$R$ is a feature structure $\sequence{Q,q,\delta,\rho}$,

$\rho$ is a total function from $Q$ to $\S$, and

for
each $\alpha\in\A$ and
each $q'\in Q$,
\disp[
\ifthen[
if
$\delta(q',\alpha)$ is defined
then
$\F(\rho(q'),\alpha)$ is defined, and
\\
$\F(\rho(q'),\alpha)\preceq\rho(\delta(q',\alpha))$.
\ifthen]
\disp]
\disp]
\prop]
Each resolved feature structure is a well-typed (see {\cite{Carpenter
1992}}) feature structure with output alphabet $\S$.

\prop[
Definition

$R$ is a resolvant of $F$ iff
\disp[
$R$ is a resolved feature structure $\sequence{Q,q,\delta,\rho}$,

$F$ is a feature structure $\sequence{Q,q,\delta,\theta}$, and

for
each $q'\in Q$,
$\theta(q')\preceq\rho(q')$.
\disp]
\prop]
\prop[
Proposition
\label{F_is_true_of_u_iff_some_resolvant_of_F_is_true_of_u}

For
each interpretation $I$,
each object $u$ in $I$ and
each feature structure $F$,
\disp[
\ifandonlyif[
$F$ is true of $u$ under $I$
iff
some resolvant of $F$ is true of $u$ under $I$.
\ifandonlyif]
\disp]
\prop]

\prop[
Definition

$\sequence{\Q,\T,\preceq,\S,\A,\F}$ is rational iff
for
each $\sigma\in\S$ and
each $\alpha\in\A$,
\disp[
\ifthen[
if
$\F(\sigma,\alpha)$ is defined
then
for
some $\sigma'\in\S$,
$\F(\sigma,\alpha)\preceq\sigma'$.
\ifthen]
\disp]
\prop]
\prop[
Proposition
\label{each_resolved_feature_structure_is_satisfiable}

If
$\sequence{\Q,\T,\preceq,\S,\A,\F}$ is rational
then
for
each resolved feature structure $R$,
$R$ is satisfiable.
\prop]
\proof[
Suppose that $R=\sequence{Q,q,\delta,\rho}$ and $\beta$ is a bijection
from ordinal $\zeta$ to $\S$.  Let
\disp[
$\Delta_0=
\left\{
\pi
\left|\begin{tabular}{l}
for some $q'\in Q$,
\\
\hspace{4pt}$\pi$ runs to $q'$ in $R$
\end{tabular}\right.\right\}$,

$\Gamma_0=
\left\{
\sequence{\pi_1,\pi_2}
\left|\begin{tabular}{l}
for some $q'\in Q$,
\\
\hspace{4pt}$\pi_1$ runs to $q'$ in $R$, and
\\
\hspace{4pt}$\pi_2$ runs to $q'$ in $R$
\end{tabular}\right.\right\}$, and

$\Lambda_0=
\left\{
\sequence{\pi,\sigma}
\left|\begin{tabular}{l}
for some $q'\in Q$,
\\
\hspace{4pt}$\pi$ runs to $q'$ in $R$, and
\\
\hspace{4pt}$\sigma=\rho(q')$
\end{tabular}\right.\right\}$.
\disp]
For each $n\in\N$, let
\disp[
$\Delta_{n+1}=
\Delta_n
\cup
\left\{
\pi\alpha
\left|\begin{tabular}{l}
$\alpha\in\A$,
\\
$\pi\in\Delta_n$, and
\\
$\F(\Lambda_n(\pi),\alpha)$ is defined
\end{tabular}\right.\right\}$,

$\Gamma_{n+1}=
\Gamma_n
\cup
\left\{
\sequence{\pi_1\alpha,\pi_2\alpha}
\left|\begin{tabular}{l}
$\alpha\in\A$,
\\
$\pi_1\alpha\in\Delta_{n+1}$,
\\
$\pi_2\alpha\in\Delta_{n+1}$, and
\\
$\sequence{\pi_1,\pi_2}\in\Gamma_n$,
\end{tabular}\right.\right\}$, \hfill and

$\Lambda_{n+1}=
\Lambda_n
\cup
\left\{
\sequence{\pi\alpha,\beta(\xi)}
\left|\begin{tabular}{l}
$\alpha\in\A$,
\\
$\pi\in\Delta_n$,
\\
$\pi\alpha\in\Delta_{n+1}\setminus\Delta_n$, and
\\
$\xi$ is the least ordinal
\\
\hspace{4pt}in $\zeta$ such that
\\
\hspace{4pt}$\F(\Lambda_n(\pi),\alpha)\preceq\beta(\xi)$
\end{tabular}\right.\right\}$.
\disp]
For each $n\in\N$, $\sequence{\Delta_n,\Gamma_n,\Lambda_n}$ is a
semi-morph.  Let
\disp[
$\Delta=\bigcup\{\Delta_n\mid n\in\N\}$,

$\Gamma=\bigcup\{\Gamma_n\mid n\in\N\}$, and

$\Lambda=\bigcup\{\Lambda_n\mid n\in\N\}$.
\disp]
$\sequence{\Delta,\Gamma,\Lambda}$ is a morph that $R$ approximates.
By theorem~\ref{F_is_satisfiable_iff_F_approximates_some_morph}, $R$
is satisfiable.
\proof]

\prop[
Theorem
\label{F_is_satisfiable_iff_F_has_a_resolvant}

If
$\sequence{\Q,\T,\preceq,\S,\A,\F}$ is rational
then
for each feature structure $F$,
\disp[
$F$ is satisfiable
iff
$F$ has a resolvant.
\disp]
\prop]
\proof[
{}From
propositions~\ref{F_is_true_of_u_iff_some_resolvant_of_F_is_true_of_u}
and~\ref{each_resolved_feature_structure_is_satisfiable}.
\proof]

\subsection*{\raggedright 5. A SATISFIABILITY ALGORITHM}

In this section, I use
theorem~\ref{F_is_satisfiable_iff_F_has_a_resolvant} to show how --
given a rational signature that meets reasonable computational
conditions -- to construct an effective algorithm to decide if a
feature structure is satisfiable.

\pagebreak

\prop[
Definition

$\sequence{\Q,\T,\preceq,\S,\A,\F}$ is computable iff
\disp[
$\Q$, $\T$ and $\A$ are countable,

$\S$ is finite,

for some effective function $\Sub$,
\disp[
for
each $\tau_1\in\T$ and
each $\tau_2\in\T$,
\disp[
\ifthenotherwise[
if
$\tau_1\preceq\tau_2$
then
$\Sub(\tau_1,\tau_2)=$ `true'
otherwise
$\Sub(\tau_1,\tau_2)=$ `false', and
\ifthenotherwise]
\disp]
\disp]

for some effective function $\App$,
\disp[
for
each $\tau\in\T$ and
each $\alpha\in\A$,
\disp[
\ifthenotherwise[
if
$\F(\tau,\alpha)$ is defined
then
$\App(\tau,\alpha)=\F(\tau,\alpha)$
otherwise
$\App(\tau,\alpha)=$`undefined'.
\ifthenotherwise]
\disp]
\disp]
\disp]
\prop]
\prop[
Proposition
\label{Resolve}

If
$\sequence{\Q,\T,\preceq,\S,\A,\F}$ is computable
then
for
some effective function $\Res$,
\disp[
for
each feature structure $F$,
\disp[
$\Res(F)=$ a list of the resolvants of $F$.
\disp]
\disp]
\prop]
\proof[
Since $\sequence{\Q,\T,\preceq,\S,\A,\F}$ is computable, for some
effective function $\Gen$,
\disp[
for
each finite $Q\subseteq\Q$,
\disp[
$\Gen(Q)=$ a list of the total functions from $Q$ to $\S$,
\disp]
\disp]
for some effective function $\Test_1$,
\disp[
for
each finite set $Q$,
each finite partial function $\delta$ from the Cartesian product of
$Q$ and $\A$ to $Q$, and
each total function $\theta$ from $Q$ to $\T$,
\disp[
\ifthenotherwise[
if
for
each $\sequence{q,\alpha}$ in the domain of $\delta$,
\disp[
$\F(\theta(q),\alpha)$ is defined, and
\\
$\F(\theta(q),\alpha)\preceq\theta(\delta(q,\alpha))$
\disp]
then
$\Test_1(\delta,\theta)=$ `true'
otherwise
$\Test_1(\delta,\theta)=$ `false',
\ifthenotherwise]
\disp]
\disp]
and for some effective function $\Test_2$,
\disp[
for
each finite set $Q$,
each total function $\theta_1$ from $Q$ to $\T$ and
each total function $\theta_2$ from $Q$ to $\T$,
\disp[
\ifthenotherwise[
if
for
each $q\in Q$,
$\theta_1(q)\preceq\theta_2(q)$
then
$\Test_2(\theta_1,\theta_2)=$ `true'
otherwise
$\Test_2(\theta_1,\theta_2)=$ `false'.
\ifthenotherwise]
\disp]
\disp]
Construct $\Res$ as follows:
\disp[
for
each feature structure $\sequence{Q,q,\delta,\theta}$,
\disp[
set
$\Listin=\Gen(Q)$ and
$\Listout=\sequence{ }$

while
$\Listin=\sequence{\rho,\rho_1,\ldots,\rho_i}$  is not empty

\ind[do
\ind=
set
$\Listin=\sequence{\rho_1,\ldots,\rho_i}$

\ifthen[
if
$\Test_1(\delta,\rho)=$ `true',
\\
$\Test_2(\theta,\rho)=$ `true', and
\\
$\Listout=\sequence{\rho_1',\ldots,\rho_j'}$
then
set
$\Listout=\sequence{\rho,\rho_1',\ldots,\rho_j'}$
\ifthen]
\ind]

\ifthen[
if
$\Listout=\sequence{\rho_1,\ldots,\rho_n}$
then
output $\sequence{\sequence{Q,q,\delta,\rho_1},\!...,
\sequence{Q,q,\delta,\rho_n}}$. 
\ifthen]
\disp]
\disp]
$\Res$ is an effective algorithm, and
\disp[
for
each feature structure $F$,
\disp[
$\Res(F)=$ a list of the resolvants of $F$.
\disp]
\disp]
\proof]

\prop[
Theorem

If
$\sequence{\Q,\T,\preceq,\S,\A,\F}$ is rational and computable
then
for
some effective function $\Sat$,
\disp[
for each feature structure $F$,
\disp[
\ifthenotherwise[
if
$F$ is satisfiable
then
$\Sat(F)=$ `true
otherwise
$\Sat(F)=$ `false'.
\ifthenotherwise]
\disp]
\disp]
\prop]
\proof[
{}From theorem~\ref{F_is_satisfiable_iff_F_has_a_resolvant} and
proposition~\ref{Resolve}.
\proof]

Gerdemann and G\"otz's Troll system (see {\cite{G\"otz 1993}},
{\cite{Gerdemann and King 1994}} and {\cite{Gerdemann (fc)}}) employs
an efficient refinement of $\Res$ to test the satisfiability of
feature structures.  In fact, Troll represents each feature structure
as a disjunction of the resolvants of the feature structure.  Loosely
speaking, the resolvants of a feature structure have the same
underlying finite state automaton as the feature structure, and differ
only in their output function.  Troll exploits this property to
represent each feature structure as a finite state automaton and a set
of output functions.  The Troll unifier is closed on these
representations.  Thus, though $\Res$ is computationally expensive,
Troll uses $\Res$ only during compilation, never during run time.

\subsection*{\raggedright References}

\refs[
\item[\cite{Carpenter 1992}]
Robert Carpenter {\sl The logic of typed feature structures}.
Cambridge tracts in theoretical computer science 32.  Cambridge
University Press, Cambridge, England.  1992.

\item[\cite{Gerdemann (fc)}]
Dale Gerdemann.  {\sl Troll: type resolution system, user's guide}.
Son\-derforschungsbereich 340 technical report.
Eberhard-Karls-Universit\"at, T\"ubingen, Germany.  Forthcoming.

\item[\cite{Gerdemann and King (1994)}]
Dale Ger\-demann and Paul John King.  {\sl The correct and efficient
implementation of appropriateness specifications for typed feature
structures}.  In these proceedings.

\item[\cite{G\"otz 1993}]
Thilo G\"otz.  {\sl A normal form for typed feature structures}.
Master's thesis.  Eberhard-Karls-Universit\"at, T\"ubingen, Germany.
1993.

\item[\cite{King 1989}]
Paul John King.  {\sl A logical formalism for head-driven phrase
structure grammar}.  Doctoral thesis.  The University of Manchester,
Manchester, England.  1989.

\item[\cite{Moore 1956}]
E.~F.~Moore.  `Gedanken experiments on sequential machines'.  In {\sl
Automata Studies}.  Princeton University Press, Princeton, New Jersey,
USA.  1956.

\pagebreak

\item[\cite{Moshier 1988}]
Michael Andrew Moshier.  {\sl Extensions to unification grammar for
the description of programming languages}.  Doctoral thesis.  The
University of Michigan, Ann Arbor, Michigan, USA.  1988.
\refs]
\end{document}